\documentclass{article}
\pdfoutput=1
\PassOptionsToPackage{numbers, compress}{natbib}
\usepackage[preprint]{neurips_2019}

\usepackage[utf8]{inputenc} % allow utf-8 input
\usepackage[T1]{fontenc}    % use 8-bit T1 fonts
\usepackage{url}            % simple URL typesetting
\usepackage{booktabs}       % professional-quality tables
\usepackage{amsfonts}       % blackboard math symbols
\usepackage{nicefrac}       % compact symbols for 1/2, etc.
\usepackage{microtype}      % microtypography

\usepackage{times}
\usepackage{epsfig}
\usepackage{graphicx}
\usepackage{amsmath}
\usepackage{amssymb}
\usepackage[usenames, dvipsnames]{color}
\usepackage[ruled]{algorithm2e}
\usepackage[noend]{algpseudocode}
\usepackage{wrapfig}
\usepackage{tablefootnote}
\usepackage{paralist}  % for compactlist
\usepackage{subfig}
\usepackage[export]{adjustbox}
\usepackage{array}
\usepackage{pbox}
\usepackage{multirow}
\usepackage{booktabs}
\usepackage{soul}
\usepackage{enumitem}
\usepackage{gensymb}
\usepackage[pagebackref=true,breaklinks=true,colorlinks=true,citecolor=blue,bookmarks=false]{hyperref}

\newcommand{\rpm}{\raisebox{.2ex}{$\scriptstyle\pm$}}
\makeatletter
\DeclareRobustCommand\onedot{\futurelet\@let@token\@onedot}
\def\@onedot{\ifx\@let@token.\else.\null\fi\xspace}

\def\eg{\emph{e.g}\onedot} 
\def\ie{\emph{i.e}\onedot} 
 
\def\etc{\emph{etc}\onedot} 
 
\def\etal{\emph{et al}\onedot}
\makeatother

\title{Boosting Supervision with Self-Supervision for Few-shot Learning}

\author{%
Jong-Chyi Su \\
UMass Amherst \\
\texttt{jcsu@cs.umass.edu}\\
\And
Subhransu Maji \\
UMass Amherst \\
\texttt{smaji@cs.umass.edu}
\And 
Bharath Hariharan \\
Cornell University \\
\texttt{bharathh@cs.cornell.edu}
}
\begin{document}
\maketitle

\begin{abstract}
  We present a technique to improve the transferability of deep
  representations learned on small labeled datasets by introducing
  self-supervised tasks as auxiliary loss functions.
  While recent approaches for self-supervised learning have shown the
  benefits of training on large unlabeled datasets, we find improvements in generalization even on small
  datasets and when combined with strong supervision.
  Learning representations with self-supervised losses reduces the
  relative error rate
  of a state-of-the-art meta-learner by 5-25\% on several few-shot
  learning benchmarks, as well as off-the-shelf deep networks on standard classification tasks when training from scratch.
  We find the benefits of self-supervision increase with the difficulty of
  the task. 
%   are observed even when the number of images in the dataset are relatively small, as few as a thousand.
  Our approach utilizes the images within the dataset to construct self-supervised losses and hence is an effective way of learning transferable representations without relying on any external training data.
\end{abstract}

\section{Introduction}
We humans can quickly learn new concepts from limited training data, but 
current machine learning algorithms cannot.
We are able to do this by relying on our past ``visual experience''.
Recent work attempts to emulate this ``visual experience'' by training a feature
representation to classify a training dataset well, with the hope that
the resulting representation generalizes not just to unseen examples
of the same classes but to novel classes as well.
This has indeed been the case when ``deep representations'' learned on
massive image classification datasets are applied to novel image
classification tasks on related domains.
This intuition also underlies recent work on few-shot learning or meta-learning.
This approach has two related shortcomings:
Because we are training the feature representation to classify a
training dataset, it may end up discarding information that might be
useful for classifying unseen examples or novel classes.
Meanwhile, the \emph{images} of these base classes themselves contain a
wealth of useful information about the domain that is discarded.

Learning from images alone without any class labels falls under the umbrella of \emph{self-supervised learning}.
The key idea is to learn about statistical regularities (\eg, likely
spatial relationship between patches, likely orientations between
images) that might be a cue to semantics.
However, these techniques require enormous amounts of data to match
\emph{fully-supervised} approaches trained on large datasets.
It is still unclear whether self-supervised learning can help in the \emph{low data} regime. 
In particular, can these techniques help prevent overfitting to training datasets and improve performance on new examples and novel classes?

We answer this question in the affirmative.
We experiment both with standard supervised training  on small labeled datasets, as well as few-shot transfer to novel classes.
We show that, with \emph{no additional training data}, adding the
self-supervised task as an auxiliary task significantly improves
accuracy in both settings and across datasets.
In particular adding self-supervised losses leads to 5-25\% relative reduction
in classification error rate of a state-of-the-art
meta-learner~\cite{snell2017prototypical} on several image
classification datasets (see Section~\ref{sec:exp}).
We observe that the benefits are greater when the task is
more challenging, such as classification among more classes, or from
greyscale or low resolution images.
On standard classification too, we find that off-the-shelf networks
trained from scratch on these datasets have lower error rate when
trained with self-supervised losses.
We conclude that self-supervision as an auxiliary task is beneficial
for learning generalizable feature representations.

% Old flowy stuff superseded by Bharath's concise stuff
%
%While deep networks trained on large labelled datasets have yeilded
%general purpose representations for many computer vision tasks, there
%are many tasks where generalization is poor.
%This paper addresses the problem of learning transferable
%visual representations in the few-shot setting;
%One is provided with a labelled dataset of images on which
%representations are to learned, which are evaluated on novel tasks on
%the same domain provided with a few
%labelled examples (e.g., different species of birds.)
%This captures the heavy-tailed nature of many practical computer
%vision tasks where a large number of labelled images may be
%available for a few categories while the vast number of categories have few
%examples.

\section{Related work}
Our work is related to a number of recent approaches for learning robust visual representations, specifically few-shot learning and self-supervised learning. Most self-supervised learning works investigate if the features from pre-training on a large unlabeled dataset can be transferred, while our work focuses on the benefit of using self-supervision as an auxiliary task in the low training data regime. 

\paragraph{Few-shot learning}
Few-shot learning aims to learn representations that generalize well to the novel classes where only few images are available.
To this end many meta-learning methods have been proposed that
\emph{simulate} learning and evaluation of representations using a
\emph{base learner} by sampling many few-shot tasks.
These include optimization-based base learners (\eg, MAML\cite{finn2017model}, gradient unrolling~\cite{ravi2016optimization}, closed form solvers~\cite{bertinetto2018meta}, convex learners~\cite{lee2019meta}, \etc). 
Others such as matching networks~\cite{vinyals2016matching} use a
nearest neighbor classifier, protoypical networks (ProtoNet)~\cite{snell2017prototypical} use
a nearest class mean classifier~\cite{mensink2013distance}, and the
approach of Garcia and Bruna~\cite{garcia2017few} uses a graph neural network to model
higher-order relationships in data.
Another class of techniques (\eg,~\cite{gidaris2018dynamic,qi2018low,qiao2018few}) model the
mapping between training examples and classifier weights
using a feed-forward network.
On standard benchmarks for few-shot learning (\eg, miniImageNet and
CIFAR) these techniques have shown better generalization than 
simply training a network for classification on the base classes.

While the literature is vast and growing, a recent
study~\cite{chen2019closerfewshot} has shown that
meta-learning is less effective when domain shifts between training
and novel classes are larger, and that the differences between
meta-learning approaches are diminished when deeper network architectures are used.
We build our experiments on top of this work and show that auxiliary self-supervised tasks 
provide additional benefits on realistic datasets and state-of-the-art
deep networks (Section~\ref{sec:exp}).

%\paragraph{Data augmentation and hallucination}
%While image-based data augmentation (\eg, scaling, jittering, \etc) are widely used for training deep networks, recent approaches employ complex strategies to generalize representations to novel classes in the few-shot setting.
%Data hallucination~\cite{hariharan2017low,WangCVPR2018a} generate novel examples based on estimated relationships between clusters of images features.
% \todo{how exactly does this work?}.
%Others such as Antoniou~\etal~\cite{antoniou2017data} use generative adversarial networks (GANs) to generate face data, while Schwartz~\etal~\cite{schwartz2018delta} generate examples based on a model of intra-class variance learned from pairs of images within a class.
%While the success of these methods have been limited the techniques are complementary to our approach and may lead to additional benefits. For simplicity we only employ image-based data augmentation strategies such as cropping and scaling (Section~\ref{sec:exp}).
% \todo{What are the drawbacks?}

\paragraph{Self-supervised learning}
% A practical benefit of this is that the images within the dataset are used which avoids the need to acquire a large dataset of images within the same domain.
Human labels are expensive to collect and hard to scale up. To this end, there has been increasing research interest to investigate learning representations from unlabeled data. In particular, the image itself already contains structural information which can be utilized. Self-supervised learning approaches attempt to capture this.

There is a rich line of work on self-supervised learning.
One class of methods removes part of the visual data (\eg, color information) and tasks the network with predicting what has been removed from the rest (\eg, greyscale images) in a discriminative manner~\cite{larsson2016learning,zhang2016colorful,zhang2017split}.
Doersch~\etal proposed to predict the relative position of patches cropped from images~\cite{doersch2015unsupervised}. 
Wang~\etal used the similarity of patches obtained from tracking as a self-supervised task~~\cite{wang2015unsupervised} and combined position and similarity predictions of patches~\cite{wang2017transitive}.
Other tasks include predicting noise~\cite{bojanowski2017unsupervised}, clusters~\cite{caron2018deep,wu2018unsupervised},
%\hj{just a comment: each image is considered as a cluster in~\cite{wu2018unsupervised}}, 
count~\cite{noroozi2017representation}, missing patches (\ie, inpainting)~\cite{pathak2016context}, motion segmentation labels~\cite{pathak2017learning}, \etc.
Doersch~\etal~\cite{doersch2017multi} proposed to jointly train four
different self-supervised tasks and found it to be beneficial. 
Recent
works~\cite{goyal2019scaling,kolesnikov2019revisiting} have compared
various self-supervised learning tasks at 
scale and concluded that
solving jigsaw puzzles and predicting image rotations are among the most effective, motivating the choice of tasks in our experiments.

The focus of most prior works on self-supervised learning is to supplant traditional fully supervised representation learning with unsupervised learning 
on large unlabeled datasets for downstream tasks.
In contrast, our work focuses
on few-shot classification setting and shows that self-supervision
helps in the low training data regime without relying on any external dataset.
% for DA~\cite{carlucci2019domain}
% \todo{Combine the two paragraphs into a single coherent one.}

\paragraph{Multi-task learning}
Our work is related to multi-task learning, a class of techniques
that train on multiple task objectives together to improve each one.
Previous works in the computer vision literature have shown moderate
benefits by combining tasks such as edge, normal, and 
saliency estimation for images, or part segmentation and
detection for humans~\cite{kokkinos2017ubernet,maninis2019attentive,ren2018cross}.
Unfortunately, in many cases, tasks can be detrimental to others resulting in reduced performance.
Moreover, acquiring additional task labels is expensive.
In contrast, self-supervised tasks do not require additional labeling
effort and we find that they often improve the generalization of learned representations.

% \todo{SM: I'm not sure how related these methods are. Perhaps move
%   these to where you describe how you balanced the two methods?}.
% Several methods are proposed to stabilize multi-task
% training, \eg balancing between the losses by normalizing the
% gradient~\cite{chen2017gradnorm} or by uncertainty
% measure~\cite{kendall2018multi}. Other methods include using
% multi-object optimization ~\cite{sener2018multi} and using different
% network routes for each task~\cite{maninis2019attentive}.
% Different from abovementioned works, our method combines supervised and unsupervised objectives, and can be trained by simply combining the objectives.

\section{Method}
\begin{figure}
\centering
\includegraphics[width=\linewidth]{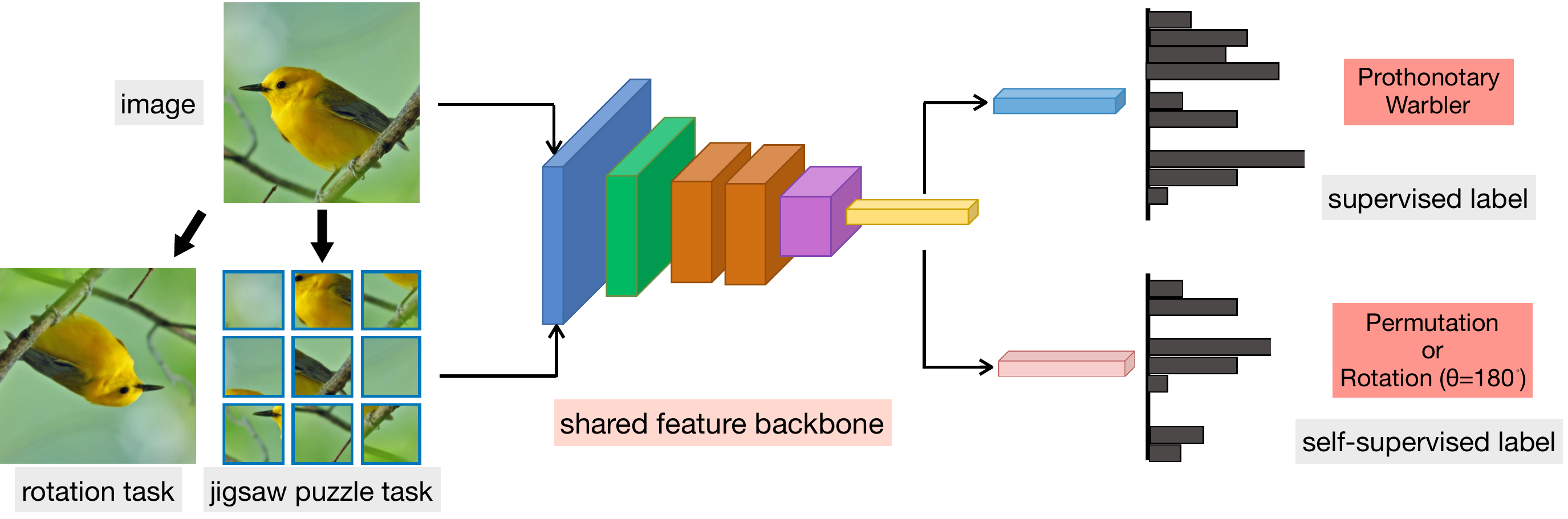}
\caption{\label{fig:overview} \textbf{Overview of our approach.} Our
  framework combined supervised losses derived from class labels and
  self-supervised losses derived from the images in the dataset to
  learn a feature representation. Since the feature backbone is
  shared, the self-supervised tasks act as an additional
  regularizer. In our experiments, the features are learned either in
  a meta-learning framework for few-shot transfer or for standard
  classification (shown above) with jigsaw puzzle or rotation tasks as
self-supervision.}
\end{figure}

Our framework augments standard supervised losses with those derived
from self-supervised tasks to improve representation learning as seen
in Figure~\ref{fig:overview}.
We are given a training dataset $\{(x_i, y_i)\}_{i=1}^n$ consisting of pairs
of images $x_i \in {\cal X}$ and labels $y_i \in {\cal Y}$.
A feed-forward CNN $f(x)$ 
maps the input to an embedding space, which is then mapped to
the label space using a classifier $g$.
The overall prediction function from the input to the labels can be written as $g \circ f (x): {\cal X}
\rightarrow {\cal Y}$.
Learning consists of estimating functions $f$ and $g$ that minimize a
loss $\ell$ on the predicted labels on the training data, along with
suitable regularization ${\cal R}$ over the functions $f$ and $g$, and
can be written as:
\[
 {\cal L}_s := \sum_i \ell\big(g\circ f (x_i), y_i\big) + {\cal R}(f, g).
\]

For example, for classification tasks a commonly used loss is the
cross-entropy loss (softmax loss) over the labels, and a suitable
regularizer is the L2 norm of the parameters of the functions.
For transfer learning $g$ is discarded and relearned on
training data for a new task. The combination can be further fine-tuned if necessary.

In addition to supervised losses, we consider self-supervised losses ${\cal L}_{ss}$
based on labeled data $x \rightarrow (\hat{x},\hat{y}$)
that can be systematically derived
from inputs $x$ alone.
Figure~\ref{fig:overview} shows two examples: the \emph{jigsaw task}
rearranges the input image and uses the index of the permutation as
the target label, while the \emph{rotation task} 
uses the angle of the rotated image as the target label. 
A separate function $h$ is used to predict these labels from the shared
feature backbone $f$ and the self-supervised loss can be written as:
\[
 {\cal L}_{ss} := \sum_i \ell\big(h\circ f (\hat{x}_i), \hat{y}_i\big).
\]
Our final loss function combines the two losses:
\[
 {\cal L} := {\cal L}_s + {\cal L}_{ss}, 
\]
and thus the self-supervised losses act as a data-dependent
regularizer for representation learning.
Below we describe various losses we consider for image classification tasks.

\subsection{Supervised losses (${\cal L}_s$)}
\label{sec:sl}
We consider two supervised loss functions depending on whether the
representation learning is aimed at \emph{standard
classification} where training and test tasks are identical, or
\emph{transfer} where the test task is different from
training. 

\emph{Standard classification loss.} On tasks we assume
that the test set consists of novel images from the same distribution
as the training set.
We use the standard cross-entropy (softmax) loss computed over posterior label
distribution predicted by the model and target label.

\emph{Few-shot transfer task loss.} Here we assume abundant labeled data on \emph{base} classes
$\mathcal{D}_b$ and limited labeled data on \emph{novel} classes $\mathcal{D}_n$ (in the terminology of~\cite{hariharan2017low}). 
We use losses commonly used in meta-learning frameworks for few-shot learning~\cite{snell2017prototypical,sung2018learning,vinyals2016matching}.
In particular we base our approach on prototypical
networks~\cite{snell2017prototypical} which perform episodic training and testing over
sampled datasets in stages called meta-training and meta-testing.
During meta-training stage, we randomly sample $N$ classes from
$\mathcal{D}_b$, then we select a support set $\mathcal{S}_b$ with $K$
images per class and another query set $\mathcal{Q}_b$ with $M$ images
per class. 
We call this an $N$-way $K$-shot classification task. 
The embeddings are trained to predict the labels of the query set
$\mathcal{Q}_b$ conditioned on the support set $\mathcal{S}_b$ using a
nearest mean (prototype) model. The objective is to minimize the
prediction loss on the query set. 
Once training is complete, given the novel dataset $\mathcal{D}_n$ class
prototypes are recomputed for classification and query examples
are classified based on the distances to the class prototypes.
The model is related to distance-based learners such as matching
networks~\cite{vinyals2016matching} or metric-learning based on label similarity~\cite{koch2015siamese}.

%While research on meta-learning algorithms for few-shot transfer is a
%very active area of research, recent work~\cite{chen2019closerfewshot}
%suggests that the differences between the approaches are smaller with
%very deep networks such as ResNet18.
%This motivated our choice of using a ResNet18 based prototypical
%network, a leading technique in~\cite{chen2019closerfewshot}, as our
%meta-learner.

%We adopt the prototypical networks (ProtoNet)~\cite{snell2017prototypical}, which performs the best on CUB with 5-shot case when using ResNet18 as backbone network.
%The ProtoNet computes the mean class features of the support set
%$\mathcal{S}_b$, then compare the distances between the query
%features and mean class features. L2 distances between features are
%used to compute the loss. On meta-testing stage, the network is fixed
%and only the class mean are computed for the query set
%$\mathcal{Q}_b$. 

\subsection{Self-supervised losses (${\cal L}_{ss}$)}
\label{sec:ssl}
We consider two losses motivated by a recent large-scale comparison of the effectiveness of various self-supervised learning tasks~\cite{goyal2019scaling}.

\emph{Jigsaw puzzle task loss.} In this case the input image $x$ is
tiled into 3$\times$3 regions and permuted randomly to obtain an input
$\hat{x}$. The target label $\hat{y}$ is the index of the permutation. 
The index (one of 9!) is reduced to one of 35 following the procedure
outlined in~\cite{noroozi2016unsupervised}, which grouped the 
possible permutations based on hamming distance as a way to control
the difficulty of the task.

\emph{Rotation task loss.} In this case the input image $x$ is
rotated by an angle $\theta \in \{0\degree,90\degree,180\degree,270\degree\}$ to obtain $\hat{x}$ and the
target label $\hat{y}$ is the
index of the angle.

In both cases we use the cross-entropy (softmax) loss between the target and prediction.

\subsection{Stochastic sampling and training}
We use the following strategy for combining the two losses for stochastic training. 
The same batch of images sampled for the supervised loss (Section~\ref{sec:sl}) are used for 
the self-supervision task. 
After the two forward passes, one for the supervised task and one for the self-supervised task,  both losses are combined for computing gradients using back-propagation.
There are several recent works proposed to stabilize multi-task training, \eg balancing the losses by normalizing the gradient~\cite{chen2017gradnorm}, by uncertainty measure~\cite{kendall2018multi}, or multi-object optimization~\cite{sener2018multi}. However, simply combining the two losses with equal weights performed well and we use this protocol for all our experiments.

%For standard classification \todo{What do we do here?}

% \todo{Perhaps describe
%   the high-level details of training set sampling and self-supervised
%   dataset sampling. How to balance the two losses. Learning issues, etc.}

%\subsection{Multi-task Learning}
%Given the two objectives, few-shot classification and auxiliary task, there are many ways to combine them. We found the most intuitive method, simply weighting the two objectives is effective:
%\[
%\mathcal{L} = (1-\lambda) \mathcal{L}_{cls} + \lambda \mathcal{L}_{aux}
%\]
%The only hyper-parameter is the $\lambda$ between two losses. We found $\lambda=0.5$ is robust across all the dataset by grid search. 

\section{Experiments}\label{sec:exp}
We present results for representation learning for few-shot transfer
learning (Section~\ref{sec:fsl}) and standard classification tasks
(Section~\ref{sec:fgvc}). 
We begin by describing the datasets and details of our experiments.

\paragraph{Datasets} We select a diverse set of image classification
datasets: Caltech-UCSD birds~\cite{cub}, Stanford
dogs~\cite{dogs}, Oxford flowers~\cite{flowers}, Stanford
cars~\cite{cars}, and FGVC aircrafts~\cite{aircrafts} for our
experiments.
Each of these datasets contains between 100 to 200 classes with
a few thousands of images as seen in Table~\ref{tab:stats}.
For few-shot learning we split classes into three disjoint sets: base,
validation, and novel. 
For each class all the images in the dataset are used in the corresponding set. 
The model is trained on the base set, and the validation set is used to select the model that has the best performance. 
The model is then tested on the novel set given few examples per class.
For birds, we use the same split as~\cite{chen2019closerfewshot}, where \{\emph{base, val, novel}\} sets have \{100, 50, 50\} classes respectively. 
For other datasets we use the same ratio for splitting the classes.
For standard classification, we follow the original train and test split for each dataset.%, and all the classes are included for both training and test stage. 

% \begin{figure}[h!]
% \centering
% \includegraphics[width=\linewidth]{figs/mosaic.png}
% \caption{\label{fig:mosaic} Example images of datasets.}
% \end{figure}

%-------------------------------------------------------------
\begin{table}[h!]
\renewcommand{\arraystretch}{1.2}
\renewcommand{\tabcolsep}{6pt}
\centering
\includegraphics[width=\linewidth]{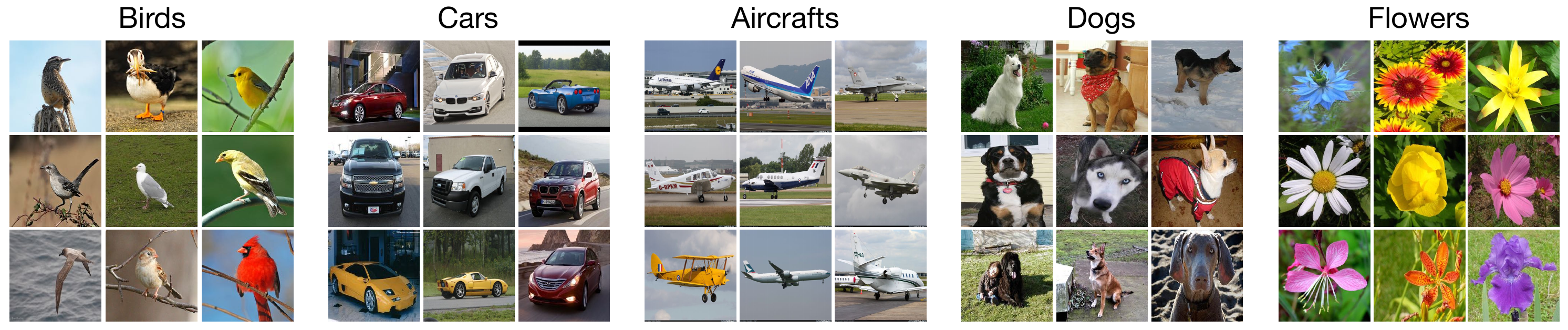}
   \begin{tabular}{c|c|c|c|c|c|c|c}
	\toprule
 		Setting & Set & Stats & Birds & Cars & Aircrafts & Dogs & Flowers \\
        \hline
		\multirow{6}{*}{Few-shot transfer} & \multirow{2}{*}{Base} & \#~class & 100 & 98 & 50 & 60 & 51 \\
		 && \#~images & 5885 & 8023 & 5000 & 10337 & 4129 \\
		 \cline{2-8}
		 &\multirow{2}{*}{Val} & \#~class & 50 & 49 & 25 & 30 & 26 \\
		 && \#~images & 2950 & 4059 & 2500 & 5128 & 2113 \\
		 \cline{2-8}
		 &\multirow{2}{*}{Novel} & \#~class &  50 & 49 & 25 & 30 & 25  \\
		 && \#~images & 2953 & 4103 & 2500 & 5115 & 1947 \\
		\hline
		\multirow{4}{*}{Standard classification} &\multirow{2}{*}{Train} & \#~class & 200 & 196 & 100 & 120 & 102  \\
		 && \#~images & 5994 & 8144 & 6667 & 12000 & 2040 \\
		 \cline{2-8}
		& \multirow{2}{*}{Test} & \#~class & 200 & 196 & 100 & 120 & 102  \\
		 && \#~images & 5794 & 8041 & 3333 & 8580 & 6149 \\
		\bottomrule
	\end{tabular}
	\vspace{0.1in}
\caption{\textbf{Example images and dataset statistics}. For few-shot experiments (top rows) the classes are split into \emph{base}, \emph{val}, and \emph{novel} set. Image representations learned on \emph{base} set are evaluated on the novel set while \emph{val} set is used for cross-validation.
For standard classification experiments we use the standard training and test splits provided in the datasets.
These datasets vary in the number of classes but are orders of magnitude smaller than ImageNet dataset.}
\label{tab:stats}
\end{table}

%-------------------------------------------------------------
\paragraph{Experimental details on few-shot learning experiments}
% For few-shot learning we choose ProtoNet~\cite{snell2017prototypical} for our model. 
% This is shown to have the best performance on CUB 5-way 5-shot classification reported in~\cite{chen2019closerfewshot}. 
We follow the best practices and codebase for few-shot learning laid
out
in Chen~\etal\cite{chen2019closerfewshot}\footnote{\url{https://sites.google.com/view/a-closer-look-at-few-shot/}}.
In particular we use a ProtoNet~\cite{snell2017prototypical} with ResNet18~\cite{he2016deep} as feature backbone.
Following~\cite{chen2019closerfewshot} we use 5-way (classes) and
5-shot (examples per-class), with 16 query images for training. 
The models are trained with ADAM~\cite{kingma2014adam} optimizer with
a learning rate of 0.001 for 40,000 episodes. 
Once training is complete (based on validation error), we report the
mean accuracy and 95\% confidence interval over 600 test experiments. 
In each test experiment, $N$ classes are selected from the novel set, and
for each class 5 support images and 16 query images are used. 
We report results for testing with $N=\{5,20\}$ classes.
During training, especially for jigsaw puzzle task, we found it to be
beneficial to \emph{not} track the running mean and variance for the batch
normalization layer, and instead estimate them for each batch
independently. 
We hypothesize that this is because the inputs contain both full-sized
images and small patches, which might have different statistics. At test
time we do the same.

\paragraph{Experimental details on standard classification experiments}
For classification we once again train a ResNet18 network from scratch. 
All the models are trained with ADAM optimizer with a learning rate of 0.001 for 400 epochs with a batch size of 16.
These parameters were found to be nearly optimal on the birds dataset
and we keep them fixed for all the other datasets.
We track the running statistics for the batch normalization layer for the softmax baselines following
the conventional setting, \ie w/o self-supervised loss, but do not track these statistics when training with
self-supervision for the aforementioned reasons. 

\paragraph{Architectures for self-supervised tasks}
The ResNet18 results in a 512-dimensional feature
vector for each input, and we add a fully-connected
(\texttt{fc}) layer with 512-units on top of this. 
For jigsaw puzzle task, we have nine patches as input, resulting in nine
512-dimensional feature vectors, which are concatenated. This
is followed by a \texttt{fc} layer, projecting the feature vector from
4608 to 4096 dimensions, and a \texttt{fc} layer with 35-dimensional outputs
corresponding to the 35 permutations for the jigsaw task.

For the rotation task, the 512-dimensional output of ResNet18 is
passed through three \texttt{fc} layers with \{512, 128, 128, 4\}
units, where the predictions correspond to the four rotation
angles. 
Between each \texttt{fc} layer, a \texttt{ReLU} (\ie,~$\max(0, x)$) activation and
a dropout layer with dropout probability of 0.5 are added.

\paragraph{Experimental details on image sampling and data augmentation}
Experiments in~\cite{chen2019closerfewshot} found that data
augmentation has large impact in the generalization performance for few-shot learning. 
We follow their procedure for all our experiments described as
follows.
For classification, images are first resized to 224 pixels for the
shorter edge while maintaining the aspect ratio, from which a central
crop of 224$\times$224 is obtained. 
For predicting rotations, we use the same cropping method then rotate
the images.
For jigsaw puzzles, we first randomly crop 255$\times$255 region from
the original image with random scaling between $[0.5,1.0]$, then split
into 3$\times$3 regions, from which a random crop of size 64$\times$64
is picked.
We use PyTorch~\cite{paszke2017automatic} for our
experiments.

%-------------------------------------------------------------
\subsection{Results on few-shot learning}\label{sec:fsl}
We first present results on few-shot learning where the models
are trained on base set then transferred and tested on a novel set. 
Table~\ref{tab:fsl_result} shows the accuracies of various models for 5-way 5-shot classification
(top half), and 20-way 5-shot classification (bottom half).
Our baseline ProtoNet (first row in each half) reproduces the results
for the birds dataset presented in~\cite{chen2019closerfewshot} (in
their Table A5). This was reported to be the top-performing method on
this dataset and others.

On 5-way classification, using jigsaw puzzles as an auxiliary task
gives the best performance, improving the ProtoNet baseline on all
five datasets. Specifically, it reduces the relative error rate by
19.6\%, 8.0\%, 4.6\%, 15.9\%, and 27.8\% on birds, cars, aircrafts,
dogs, and flowers datasets respectively. 
% 2.5\%, 0.7\%, 0.4\%, 2.7\% and 3.0\% out of 12.7, 8.3, 8.6, 17.0, 10.8 error rate which reduces
% 19.6\%, 8\%, 4.6\%, 15.9\%, 27.8\% relative error rate 
Predicting rotations also improves the ProtoNet baseline on birds, cars, and dogs.
We further test the models on the 20-way classification task, as
shown in the bottom half of the table. 
The relative improvements using self-supervision are greater in this setting. 

We also test the accuracy of the model trained only with self-supervision. 
Compared to the randomly initialized model (``None'' rows, Table~\ref{tab:fsl_result}), training the network to predict rotations gives around 2\% to 7\% improvements on all five datasets, while solving jigsaw puzzles only improves on aircrafts and flowers. 
% Note that the model with random parameters still performs better
% than chance performance.

To see if self-supervision is beneficial for harder classification
tasks, we conduct experiments on degraded images. 
For cars and aircrafts, we use low-resolution images where the images
are down-sampled by a factor of \emph{four} and up-sampled back to 224$\times$224 with
bilinear interpolation. 
For natural categories we discard color and perform experiments with
greyscale images.
Low-resolution images are considerably harder to classify for man-made
categories while color information is most useful for natural
categories\footnote{see~\cite{su2017adapting} for a discussion on how color and resolution affect fine-grained
classification results}.
The results for this setting are shown in Table~\ref{tab:low_quality_result}. 
On birds and dogs datasets, the improvements using
self-supervision are higher compared to color images. 
Similarly on the cars and aircrafts datasets with low-resolution
images, the improvement goes up from 0.7\% to 2.2\% and from
0.4\% to 2.0\% respectively.
% To this end, we conduct experiments on greyscale images, where the improvements on aircrafts increases to 1.0\%.
% On aircrafts, we speculate that since the airplanes are mostly on horizontal direction, learning features for vertical directions may not help.
% On man-made rigid objects like cars, the benefit is smaller since the color information is usually not a discriminative feature (\eg different cars can have the same color.) 

%Note that the miniImageNet dataset proposed
%by~\cite{vinyals2016matching,ravi2016optimization} includes many
%images that are too small for jigsaw puzzle loss, so we did not
%include the miniImagenet results. 

Our initial experiments on miniImageNet showed small improvements from
75.2\% of ProtoNet baseline to 75.9\% when training with the jigsaw puzzle loss.
We found that, perhaps owing to the low resolution images in this
dataset, the jigsaw puzzle task was significantly harder to train (we
observed low training accuracy for the self-supervised
task). Perhaps other forms of self-supervision might be beneficial
here.

\begin{table}[t!]
  \renewcommand{\arraystretch}{1.3}
  \setlength{\tabcolsep}{5.5pt}
  \centering
  \begin{tabular}{c|c|c|c|c|c}
    \toprule
    \multirow{2}{*}{Loss}& Birds & Cars & Aircrafts & Dogs & Flowers\\
    \cline{2-6}
    &\multicolumn{5}{c}{5-way 5-shot} \\
    \hline
    ProtoNet & 
    87.29\rpm0.48 &91.69\rpm0.43 &91.39\rpm0.40 &82.95\rpm0.55 &89.19\rpm0.56
    \\
    % 		\hline
    ProtoNet + Jigsaw & 
    \textbf{89.80\rpm0.42}&\textbf{92.43\rpm0.41}&\textbf{91.81\rpm0.38}&\textbf{85.66\rpm0.49}&\textbf{92.16\rpm0.43}
    \\
    Jigsaw & 25.73\rpm0.48&25.26\rpm0.46&38.79\rpm0.61&24.27\rpm0.45&50.53\rpm0.73%
    \\
    % 		\hline
    ProtoNet + Rotation & 89.39\rpm0.44&92.32\rpm0.41&91.38\rpm0.40&84.25\rpm0.53&88.99\rpm0.52%**
    \\
    Rotation & 33.09\rpm0.63&29.37\rpm0.53&29.54\rpm0.54&27.25\rpm0.50&49.44\rpm0.69%
    \\
    % 		\hline
    None & 26.71\rpm0.48&25.24\rpm0.46&28.10\rpm0.47&25.33\rpm0.47&42.28\rpm0.75%
    \\
    \hline
    &\multicolumn{5}{c}{20-way 5-shot}\\
    % 		\hline
    % 		Loss & CUB & Cars & Aircrafts & Dogs & Flowers\\
    \hline
    ProtoNet & 69.31\rpm0.30&78.65\rpm0.32&78.58\rpm0.25&61.62\rpm0.31&75.41\rpm0.28%
    \\
    ProtoNet + Jigsaw & 
    \textbf{73.69\rpm0.29}&79.12\rpm0.27&\textbf{79.06\rpm0.23}&\textbf{65.44\rpm0.29}&\textbf{79.16\rpm0.26}%
    \\
    Jigsaw & 8.14\rpm0.14&7.09\rpm0.12&15.44\rpm0.19&7.11\rpm0.12&25.72\rpm0.24%
    \\
    ProtoNet + Rotation & 
    72.85\rpm0.31&\textbf{80.01\rpm0.27}&78.38\rpm0.23&63.41\rpm0.30&73.87\rpm0.28%**
    \\
    Rotation & 12.85\rpm0.19&9.32\rpm0.15&9.78\rpm0.15&8.75\rpm0.14&26.32\rpm0.24%
    \\
    None & 9.29\rpm0.15&7.54\rpm0.13&8.94\rpm0.14&7.77\rpm0.13&22.55\rpm0.23%
    \\
    \bottomrule
  \end{tabular}
  \vspace{0.1in}
  \caption{\textbf{Performance on few-shot transfer task.} The mean accuracy (\%) and the 95\% confidence interval of 600 randomly chosen test experiments are reported for various combinations of loss functions. 
    The top part shows the accuracy on 5-way 5-shot classification tasks, while the bottom part shows the same on 20-way 5-shot. 
    Adding \emph{jigsaw puzzle loss} as self-supervision to the ProtoNet loss gives the best performance across all five datasets on 5-way results. 
    On 20-way classification, the improvements are even larger.
    The last row indicate results with a randomly initialized network.}
  \label{tab:fsl_result}
%   \vspace{0.1in}
\end{table}

\begin{table}[t!]
  \renewcommand{\arraystretch}{1.2}
  \setlength{\tabcolsep}{5.5pt}
  \centering
  \begin{tabular}{c|c|c|c|c|c}
    \toprule
    \multirow{2}{*}{Loss} & Birds & Cars & Aircrafts & Dogs & Flowers\\
    % 		\hline
    & Greyscale & Low-resolution & Low-resolution & Greyscale & Greyscale \\
    \hline
    &\multicolumn{5}{c}{5-way 5-shot}\\
    \hline
    ProtoNet & 82.24\rpm0.59 & 84.75\rpm0.52 & 85.03\rpm0.53 & 80.66\rpm0.59 & 86.08\rpm0.55\\
    ProtoNet + Jigsaw & 85.40\rpm0.55& 86.96\rpm 0.52 & 87.07\rpm0.47
    & 83.63\rpm0.50 & 87.59\rpm0.54\\
    \hline
    &\multicolumn{5}{c}{20-way 5-shot}\\
    \hline
    ProtoNet & 60.76\rpm0.35 & 64.72\rpm0.33 & 64.10\rpm0.27 & 57.35\rpm0.29 & 69.69\rpm0.26\\
    ProtoNet + Jigsaw & 65.73\rpm0.33 & 68.64\rpm0.33 & 68.28\rpm0.27 & 61.16\rpm0.29 & 71.64\rpm0.27\\
    \bottomrule
  \end{tabular}
  \vspace{0.1in}
  \caption{\textbf{Performance on few-shot transfer task with degraded inputs.}
    Accuracies are reported on novel set for 5-way 5-shot and 20-way
    5-shot classification
    with degraded inputs of the standard datasets, with and without \emph{jigsaw puzzle loss}. 
    The loss of color or resolution makes the task more challenging as
    seen by the drop in the performance of the baseline
    ProtoNet. However the improvements using the \emph{jigsaw puzzle
      loss} are higher in comparison to the results presented in
    Table~\ref{tab:fsl_result}.
    \label{tab:low_quality_result}}
    % \vspace{0.1in}
\end{table}
%-------------------------------------------------------------
\subsection{Results on fine-grained classification}\label{sec:fgvc}
Next, we present results using the standard classification task. 
Here we investigate if the self-supervised tasks can help
representation learning for deep networks (ResNet18 network) trained
from scratch using images and labels in the training set only.
Table~\ref{tab:fgvc} shows the accuracy using various combinations of
loss functions.
One can see that training with self-supervision improves on all
datasets we tested. 
On birds, cars, and dogs, predicting rotation gives 4.1\%, 3.1\%, and
3.0\% improvements, while on aircrafts and flowers, the jigsaw puzzle
loss gives boosts of 0.9\% and 3.6\% respectively.

\begin{table}[h!]
  \renewcommand{\arraystretch}{1.2}
  \setlength{\tabcolsep}{5.5pt}
    \centering
    \begin{tabular}{c|c|c|c|c|c}
      \toprule
      Loss & Birds & Cars & Aircrafts & Dogs & Flowers\\
      \hline
      Softmax & 47.0 & 72.6 & 69.9 & 51.4 & 72.8\\
      Softmax + Jigsaw~~~ & 49.2 & 73.2 & \textbf{70.8} & 53.5 & \textbf{76.4}\\
      Softmax + Rotation & \textbf{51.1} & \textbf{75.7} & 70.0 & \textbf{54.4} & 73.5\\
      \bottomrule
    \end{tabular}
    \vspace{0.1in}
    \caption{\textbf{Performance on standard classification task.} Per-image accuracy (\%) on the test set are reported. Using self-supervision improves the accuracy of a ResNet18 network trained from scratch over the baseline of supervised training with cross-entropy (softmax) loss on all five datasets.}
    \label{tab:fgvc}
\end{table}

% \begin{table}[t!]
% 	\renewcommand{\arraystretch}{1}
%     \setlength{\tabcolsep}{5.5pt}
%     \centering
% 	\begin{tabular}{c|c|c|c|c|c|c}
% 		\toprule
% 		\multirow{2}{*}{Loss} &\multicolumn{6}{c}{\#~training images per class}\\
% 		\cline{2-7}
% 		 & 5 & 10 & 15 & 20 & 25 & Whole training\\
%         \hline
% 		Softmax & &&&&&\\
% 		Softmax + Jigsaw & &&&&&\\
% 		Softmax + Rotation & &&&&&\\
% 		\bottomrule
% 	\end{tabular}
% 	\caption{Varying number of training images.}
% 	\label{tab:result_ablation}
% \end{table}

\subsection{Visualization of learned models}
\begin{figure}
\centering
\includegraphics[width=\linewidth]{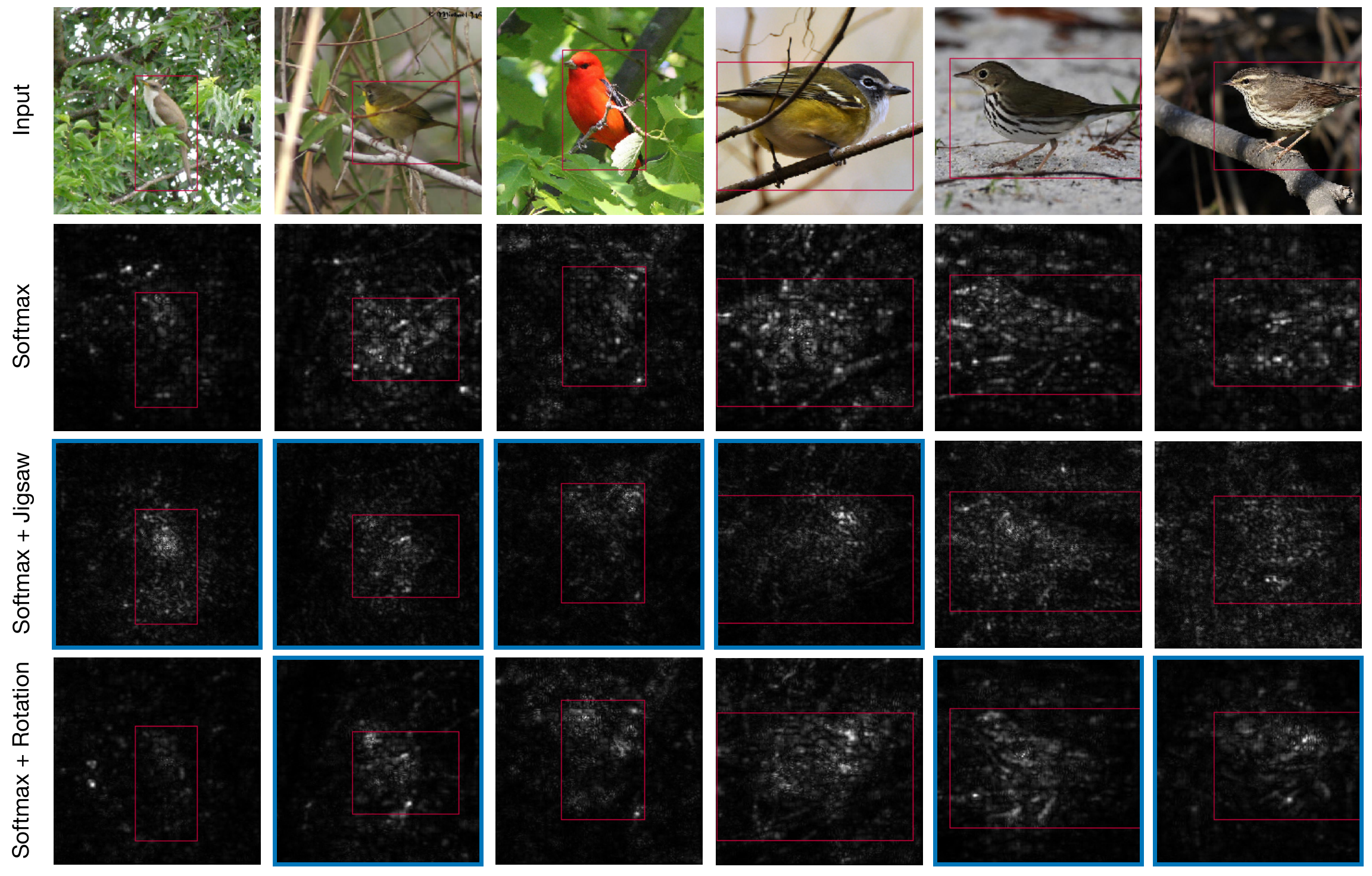}
\caption{\label{fig:saliency} \textbf{Saliency maps for various images
    and models.} For each image we visualize the magnitude of the
  gradient with respect to the correct class for models trained with
  various loss functions.
  The magnitudes are scaled to the same range for easier
  visualization.
  The models trained with self-supervision often have lower energy on
  the background regions when there is clutter. 
  We highlight a few examples with blue borders and the bounding-box of the object for each image is shown in red.
}
\end{figure}

To better understand what causes the representations to generalize
better we visualize what pixels contribute the most to the correct
classification for various models.
In particular, for each image and model we compute the gradient of the
logits (predictions before softmax) for the correct class with
respect to the input image.
The magnitude of the gradient at each pixel is a proxy for the
importance and are visualized as ``saliency maps''.
Figure~\ref{fig:saliency} shows these maps for various images and
models trained with and without self-supervision.
It appears that the self-supervised models tend to focus more on the
foreground regions, as seen by the amount of bright pixels within the
bounding box.
One hypothesis is that self-supervised tasks force the model to rely
less on background features, which might be accidentally correlated to
the class labels.
For fine-grained recognition localization indeed improves performance
when training from few examples (see~\cite{WertheimerCVPR2019} for a contemporary
evaluation of the role of localization for few-shot learning).

\section{Conclusion}
We showed that self-supervision improves transferability of
representations for few-shot learning tasks on a range of image
classification datasets.
Surprisingly, we found that self-supervision is beneficial
even when the number of images used for self-supervision is small,
orders of magnitude smaller than previously reported results.
This has a practical benefit that the images within small datasets can
be used for self-supervision without relying on a large-scale external dataset. 
Future work could investigate if additional unlabeled data within
the domain, or combination of various self-supervised losses can be
used to further improve generalization.
Future work could also investigate how and when self-supervision
improves generalization by analyzing transferabilty across a range of 
self-supervised and supervised tasks empirically~\cite{achille2019task2vec,zamir2018taskonomy}.

{\small
\bibliographystyle{ieee_fullname}
\bibliography{reference}
}

\end{document}